\documentclass{llncs}
\usepackage{graphicx}

\title{Pattern-Based Classification:\\
A Unifying Perspective}
\author{Bj\"orn Bringmann \and Siegfried Nijssen \and Albrecht Zimmermann }
\institute{Department of Computer Science\\
Katholieke Universiteit Leuven\\
Celestijnenlaan 200A, 3001, Leuven, Belgium}

\begin{document}

\maketitle

\begin{abstract}
The use of patterns in predictive models is a topic that has received a lot of attention
in recent years. Pattern mining can help to obtain models for structured domains, such as graphs and sequences, and has been proposed as a means to obtain more accurate and more interpretable models. Despite the large amount of publications devoted to this topic, we believe however that an overview of what has been accomplished in this area is missing. This paper presents our perspective on this evolving area. We identify the principles of pattern mining that are important when mining patterns for models and provide an overview of pattern-based classification methods. We categorize these methods along the following dimensions: (1) whether they {\em post-process} a pre-computed set of patterns or {\em iteratively} execute pattern mining algorithms; (2) whether they select patterns {\em model-independently} or whether the pattern selection is {\em guided by a model}. We summarize the results that have been obtained for each of these methods.
\end{abstract}

\section{Introduction}
Important problems in data mining and machine learning are classification and pattern mining. In recent years an increasing number of publications have studied the combination of these problems. The main idea in these methods is that patterns can be used to define features or can be used as rules; classification models which make use of these features or rules may be more accurate or more simple to understand. Last, but not least, in structured domains, pattern mining can be considered a propositionalization approach which enables the use of propositional data mining and machine learning algorithms. 

Despite the large amount of publications devoted to this topic, we believe however that an overview of what has been accomplished in this area is missing. It is not uncommon for publications in this area to refer to only a small portion of relevant related work, hence preventing deeper insight or a general theory from evolving. As an example, Kralj et al. pointed out that the problems of subgroup discovery, contrast set mining and emerging pattern mining are so similar that their main differences are arguably the terminology used \cite{KraljNovakEtAl-JMLR2009}. We believe that this phenomenon is much more wide-spread. For instance, in this paper we will point out that the independently proposed areas of correlating (or correlated) itemset mining and discriminative itemset mining are also mostly identical  to the problems studied in \cite{KraljNovakEtAl-JMLR2009}, except in name. 

The need to obtain a better insight in the accomplishments of this area has been observed by other authors. In particular, this has led to a tutorial at ICDM'07 by Bailey and Dong \cite{bailey07icdm} (and an extensive online reference list), a tutorial at ICDM'08 by Cheng et al. \cite{cheng08icdm} and a workshop at ECML PKDD \cite{knobbe08}. 
In this paper, we present our perspective on this area, which differs from earlier perspectives in several key aspects.
\begin{description}
 \item[Pattern Type Independence:] Other overviews have stressed the fact that there are different types of data, such as graph-based, tree-based and itemset-based data. They coupled pattern selection strategies to particular pattern types, and stressed the fact that different pattern mining algorithms are needed to deal with each such data type. Even though this is true, and indeed one often needs to implement a different pattern miner to deal with a pattern type at hand, we believe that it is more important in this case to stress the conceptual similarities between these pattern mining algorithms. Doing so leads to the insight that most approaches that have been proposed for complex data types, such as graphs, can easily also be implemented in pattern mining algorithms for simpler data types, such as itemsets; this leads to a large number of additional approaches that itemset-mining based
 approaches could be compared with.
 \item[Data Structure Independence:] In a similar way, other tutorials have stressed the fact that even for the same data type, such as itemset data, different data structures may be used to speed-up the computation of the patterns; examples are the FP-Trees \cite{cheng08icdm} and ZBDDs \cite{bailey07icdm}. Even though the choice for such data structures may have a significant impact on the efficiency of the computation, we believe that most pattern-based classification problems are orthogonal to the choice of such data structures: most solutions can be combined with any such data structure. 
 \item[Iterative Mining:] Initial approaches which combined pattern mining and classification models took a strict step-wise approach, in which a set of patterns is computed once and these patterns are subsequently used in models. However, in more recent years a large number of methods have been proposed which aim at integrating pattern mining, feature selection and model construction. In this paper we give a central position to such approaches.
\end{description}
In Section~\ref{overview} we present the key components of our proposed framework. The state-of-the-art of these components is discussed in more detail in subsequent sections.

\section{Overview}
\label{overview}
The main idea of pattern-based classification is that patterns define new features, which can be used in a classification model. A simple example is provided in the figure below. 
\begin{center}
 \includegraphics[width=10cm]{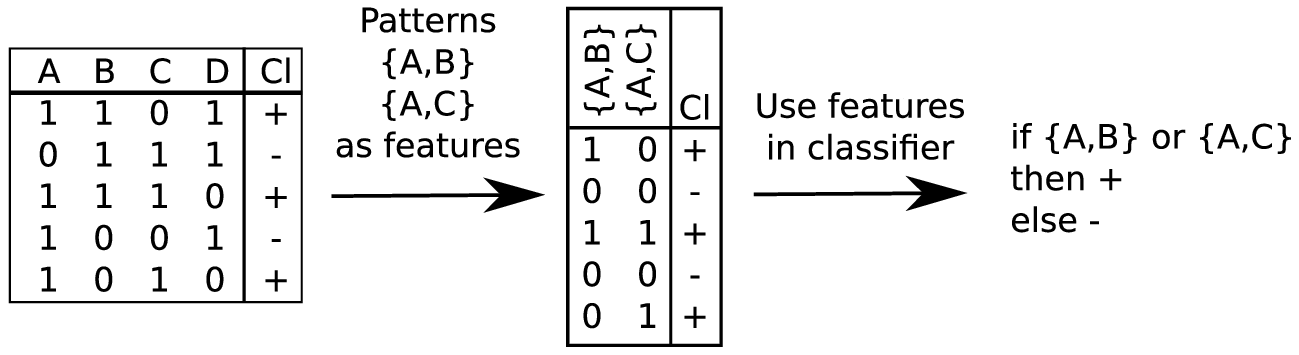}
\end{center}
Essentially, a pattern is a regularity that is observed in a number of examples, in our example $\{A,B\}$ is a pattern that occurs in the first example and third example. Whether this regularity is present or not in an example can be seen as a feature of each example. A prediction can be based on this, for instance, if an example includes items $A$ and $B$ we may predict the example to be positive.

The key challenges in finding pattern-based models are:
\begin{itemize}
 \item how to find a set of patterns; 
 \item how to combine patterns into models.
\end{itemize}
We distinguish approaches in the literature along the following dimensions:
\begin{description}
 \item[Iterative Mining or Post-Processing:] when a set of patterns is constructed, this can be done in two ways. We can run a pattern mining algorithm once to find a large set of patterns, and post-process its result to obtain a smaller set, or we can iteratively run a pattern mining algorithm, in each round finding a very small number of patterns (often only one), taking into account previous patterns in each round.
 \item[Model-Dependence or Model-Independence:] when we search a set of patterns, we can use two types of criteria. We can use criteria that \emph{explicitly} take into account the type of model in which the pattern will be used, or we can use criteria which are independent of the model; typically, in such a model-independent approach the aim is to find a set of patterns which is sufficiently diverse such that a more complex model, like an SVM, can be learned on the new features.
\end{description}
Many approaches have been developed along each of these dimensions. We will provide an overview of these approaches in Sections~\ref{independent-post-processing}, \ref{dependent-post-processing}, \ref{independent-iterative} and \ref{dependent-iterative}.
The following table clarifies how these sections correspond to these dimensions.

\begin{center}
 \begin{tabular}{rccc}
   & \textsc{Model-Dependent} & \ \ \ \  & \textsc{Model-Independent} \\ \hline
\textsc{Post-Processing} & Section~\ref{independent-post-processing} & & Section~\ref{dependent-post-processing} \\
\textsc{Iterative} & Section~\ref{independent-iterative} & & Section~\ref{dependent-iterative} \\
\end{tabular}
\end{center}

%
\begin{figure}
	\centering
	\includegraphics[width=\textwidth]{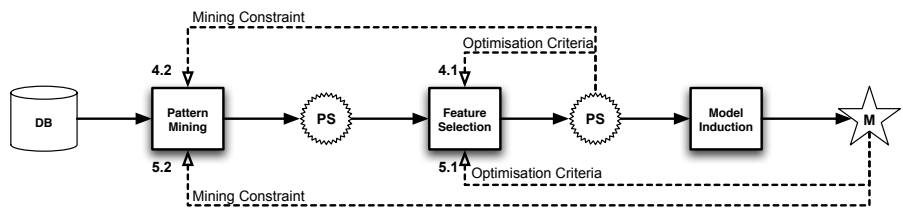}
	\vspace{-5mm}
	\caption{\label{completeFlow}The overall process from pattern mining over feature selection to model induction. The dashed arrows show four possibilities of steering the process resulting from the two different dimensions we identified.}
\end{figure}
The general search strategy for all these approaches can be summarized as in Figure~\ref{completeFlow}. Starting from a given set of examples the first step is to mine for patterns $PS$ satisfying given constraints. In a following, optional, step a subset of these patterns is selected in order to optimize this set of patterns as features in a model. Eventually those patterns are used as features to induce a model $M$ in a third step. The process allows for several ways of feedback. Each of the intermediate results can be evaluated with regard to its quality to derive constraints which can be used to guide the search/selection of further patterns or even to restart the mining or selection step with adjusted parameters. If this feedback involves explicit consultation of the induced model, we refer to the aforementioned model-dependent methods, otherwise to model-independent ones.
Most approaches in the literature can be described with this model and use one of the four different types of self-steering in the process which we will discuss in the subsequent sections.

In all these approaches, an important problem is how to find one or more patterns, iteratively or not. The simplest approach is the post-processing approach which operates on frequent patterns. However, most approaches are more sophisticated and take the class attribute into account while mining patterns. Hence, an important question in all of these approaches, iterative, model-dependent or not, is how to find patterns that take a class attribute into account. We start with an overview of solutions to this problem in Section~\ref{patternmining}.

\newcommand{\pn}{class-sensitive}
\section{Class-Sensitive Patterns}
\label{patternmining}
The starting point for taking class labels into account is in all cases to compute the (possibly weighted) support of a pattern in all classes individually. One can distinguish these approaches for using the class-specific supports in constraints:
\begin{itemize}
 \item support constraints per class, for instance, a minimum support constraint on one class combined with a maximum support constraint on another class. Such constraints can involve explicit thresholds, as for 
 {\em version space patterns} \cite{DBLP:conf/icml/KramerR01},
a minimum difference between support values for \emph{emerging patterns},
 or a maximum support of zero for an individual class as for {\em jumping emerging patterns} \cite{DBLP:journals/kais/LiDR01,DBLP:journals/tkde/FanR06}. \\
 \item constraints on scores computed from supports, sometimes in addition to support constraint. Many alternative measures for correlation strength have been proposed, ranging from \emph{confidence}, \emph{lift}, \emph{weighted relative accuracy} or \emph{novelty}, to $\chi^2$, the \emph{correlation coefficient}, \emph{information gain}, \emph{Fisher score} and others, including measures derived from classification models, such as in gBoost \cite{gboost1}. 
\end{itemize}
Patterns satisfying constraints on derived scores have been called \emph{emerging patterns} \cite{DBLP:conf/kdd/DongL99}, \emph{subgroup descriptions} \cite{kloesgen96:coll,Wrobel97:proc,DBLP:conf/pkdd/GrosskreutzRW08}, \emph{contrast sets} \cite{bay-pazzani}, \emph{correlating patterns} \cite{morishita}, \emph{discriminative patterns} \cite{cheng07}, and \emph{interesting rules} \cite{DBLP:conf/kdd/BayardoA99,DBLP:conf/vldb/MorimotoFMTY98}.
In this case one may not be interested in finding all patterns satisfying the constraints. 
Instead, one may
be interested in finding top-$k$ scoring patterns, or finding top-$k$ patterns per instance in the training data \cite{DBLP:conf/sdm/WangK05}.

It was pointed out in \cite{KraljNovakEtAl-JMLR2009} that contrast sets, emerging patterns and subgroups are compatible terms, in the sense that these terms serve the same purpose of denoting patterns that score high with respect to a scoring function that takes class labels into account. A similar observation can be made regarding correlated patterns, discriminative patterns and interesting patterns. In this paper we do not endeavour to make a choice for one of these terms; to avoid this we will call such patterns {\em \pn\ patterns} for the course of this paper. Whether the community should agree on a common name, and which one this should be, is not an issue we wish to discuss here.

In many cases threshold-based constraints are not effective enough to obtain smaller, non-redundant sets of patterns. One means to obtain smaller sets of patterns is to extend condensed representations, such as \emph{closed}, \emph{free} and \emph{non-derivable} patterns \cite{closed,free,non-derivable}, to the context of \pn\ patterns \cite{DBLP:journals/jair/Webb95,DBLP:conf/sdm/WangK05,DBLP:conf/pkdd/GarrigaKL06}.

Given the similarity in purpose of these patterns, it is not surprising that similar search strategies have been developed for each of them. Approaches that have been studied include post-processing frequent itemsets \cite{DBLP:conf/pkdd/AtzmullerP06,DBLP:journals/aai/KavsekL06,cheng07,cba} (for subgroups, discriminative patterns, interesting patterns, emerging patterns, in some cases with an additional support threshold), \emph{branch-and-bound} search \cite{DBLP:journals/jair/Webb95,Wrobel97:proc,bay99,morishita,DBLP:conf/sdm/WangK05,cheng08,DBLP:conf/kdd/ArunasalamC06,DBLP:conf/pkdd/GrosskreutzRW08,gboost1} (for subgroups, correlated patterns, contrast sets, discriminative patterns, gBoost), or variations of \emph{iterative deepening} \cite{simpler-patterns,DBLP:conf/sigmod/YanCHY08,DBLP:conf/dawak/CerfGSB08} (for correlated patterns, discriminative patterns). The reason that branch-and-bound searches have been proposed for \pn\ pattern mining is that finding such patterns has been proved to be computationally hard. Proofs can be found in \cite{morishita,DBLP:journals/tcs/WangZDL05}.

A main difference between the papers studying \pn\ patterns, is the choice for the scoring function. For instance, weighted relative accuracy is commonly used in subgroup discovery, while $\chi^2$ is common in correlated pattern mining. Insight in the differences between these measures can be obtained by comparing them in ROC space  \cite{DBLP:journals/ml/FurnkranzF05,nijssen05multi,DBLP:conf/kdd/NijssenGR09}. Among others, such studies allow to compare how well the different measures can be bounded in a branch-and-bound search. Furthermore, such studies led to the insight that for some pattern domains (such as itemsets) better bounds exist than for other pattern domains \cite{DBLP:conf/kdd/NijssenGR09}; however, \emph{all} bounds introduced before \cite{DBLP:conf/kdd/NijssenGR09} are pattern-domain independent, allowing for the application of existing strategies.

Despite that most bounds are pattern-domain independent, the combination of such bounds with approaches for dealing with particular pattern domains has received significant attention. Pattern domains that have been studied are itemset or attribute-value data (including \cite{cheng07,morishita,bay-pazzani,DBLP:conf/kdd/BayardoA99,DBLP:journals/tkde/FanR06}), sequences \cite{simpler-patterns,DBLP:journals/tcs/HiraoHSTA03}, tree-structured patterns \cite{ctc,hashimoto-sig-trees-sugar}, and graphs (including \cite{simpler-patterns,dt-gbi,DBLP:conf/sigmod/YanCHY08,gboost1}). Approaches for \pn\ itemset mining have been implemented using optimized data structures such as FP-trees \cite{Han-fpgrowth,cheng08,DBLP:conf/pkdd/AtzmullerP06} or binary decision diagrams (BDDs) \cite{DBLP:conf/kdd/LoekitoB06}.
In the remainder of this paper, we will present approaches independent of the data type or data structure for which they were proposed, as most approaches are conceptually independent of this.

An issue which has received limited attention is that of false positives. It is likely that in an exhaustive branch-and-bound search a pattern with a high score can be found, but this pattern may overfit the training data. How to control this error seems to be an open question; initial approaches suggest for instance to modify the scoring function \cite{bay-pazzani,DBLP:journals/ml/Webb07}.


\section{Model-Independent Pattern Selection}


Mining \pn\ patterns is usually the easy part, however. In many settings the number of patterns that is found is too large, in the sense that building classifiers on them is inefficient, overfitting is likely, and interpretability of the models may be hard. For patterns to be actually useful, there is the need to create a more compact set of effective patterns. If we do no take the subsequent model into account, the main aim of the {\em pattern selection} step is to reduce the redundancy of the pattern set. We can distinguish approaches which achieve this by post-processing an initial set of patterns, and approaches which iteratively search for patterns that increase the diversity of the pattern set.

\subsection{\label{independent-post-processing}Model-independent post-processing}
\begin{figure}
	\vspace{-5mm}
	\centering
	\includegraphics[width=0.75\textwidth]{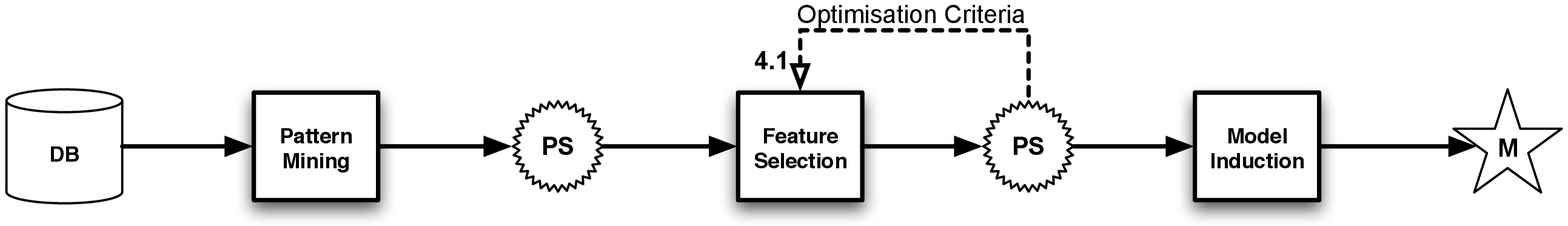}
	\vspace{-5mm}
	\caption{\label{flow_MI-PP}The model-independent post-processing approach: feedback is given by evaluation of the partially selected subset to steer feature selection.}
	\vspace{-5mm}
\end{figure}

Post-processing the result of pattern mining according to certain criteria has two distinct advantages: first, both for finding the initial set of patterns as for reducing this set, we can use or adapt existing well-developed pattern mining techniques, which are usually rather efficient. Second, it is in many cases possible to explicitly control the properties of the resulting pattern set or at least to give guarantees about them.

A variety of measures and constraints, and algorithms finding sets that satisfy them, have been proposed so far. In many cases, one is interested in finding a set of patterns that optimize a global criterion of diversity based on the occurrences of patterns in the data, sometimes in addition to explicit constraints. An example of a global criterion is {\em entropy}: if we select $n$ patterns, we can encode every example in the data with a bit-vector of length $n$. This gives every bit-vector of length $n$ a probability  in the data. The entropy of this distribution can be used as a measure of diversity.
Such sets of diverse sets can be searched exhaustively \cite{DBLP:conf/kdd/KnobbeH06,patternteams,DBLP:conf/sdm/RaedtZ07}. In practice, these approaches do not scale well, and more greedy search strategies are needed. While the focus often is somewhat different, the general technique for selecting a subset of patterns by post-processing is very similar to filter approaches for \emph{feature selection}.

Initial proposals for measures of diversity did not provide for approximation guarantees that the pattern sets found were provably good \cite{DBLP:conf/dis/DongZWL98}. However, more recently several pattern set criteria have been shown to be submodular, and consequently a greedy hill-climbing algorithm, which iteratively adds a highest scoring pattern to an initially empty set, achieves a result which approximates the optimum \cite{DBLP:conf/icdm/GarrigaHS07,thoma09}. Other recent approaches attempt to reduce the computational complexity of the pattern selection further, and study different measures for selecting patterns \cite{DBLP:conf/dis/DongZWL98,chosen-few}.

An alternative to data-only approaches is also to take into account the similarity between the pattern structures, hence taking into account mutual similarities between patterns. Also here one can define optimization criteria and greedy approaches for optimizing them  \cite{DBLP:conf/icdm/HasanCSBZ07,DBLP:conf/kdd/XinCYH06}.

A third set of approaches does not optimize a measure of diversity directly, but rather aims at finding a compact representation of the data; the idea is here that we wish to find a small set of patterns which allows to encode transactions as accurately as possible with as few patterns as possible. One can distinguish the MDL based approaches here \cite{DBLP:conf/sdm/SiebesVL06,DBLP:conf/pkdd/LeeuwenVS06}, as well as the discrete basis problem \cite{DBLP:journals/tkde/MiettinenMGDM08,DBLP:conf/pkdd/Miettinen08}.

Finally, machine learning-inspired sampling and verification techniques may also be used to obtain more diverse sets of patterns \cite{aggr-subset}.

\subsection{\label{independent-iterative}Model-independent iterative mining}
\begin{figure}
	\vspace{-5mm}
	\centering
	\includegraphics[width=0.75\textwidth]{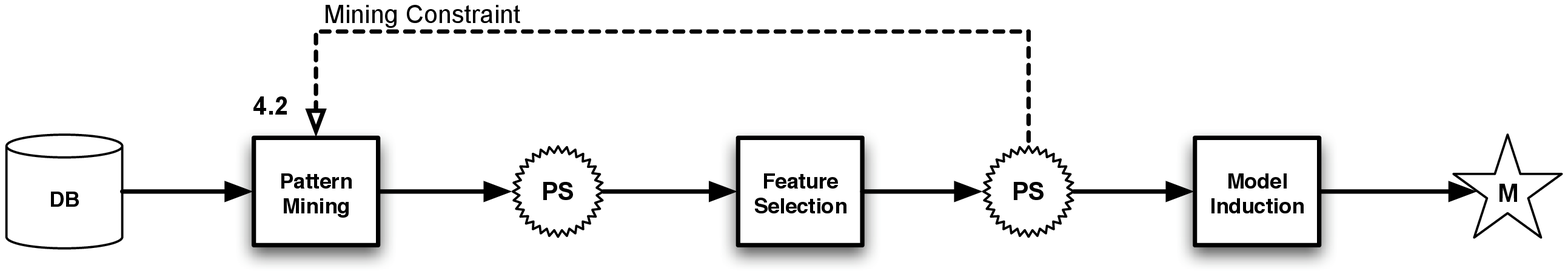}
	\vspace{-5mm}
	\caption{\label{flow_MI-IT}The model-independent iterative approach: the feedback derived by evaluating the pattern set directly influences the mining of individual patterns.}
	\vspace{-5mm}
\end{figure}
The alternative to feature selection lies in \emph{feature construction}. In general, the idea is here to avoid generating patterns beforehand, but to search for patterns during the selection process.
The main advantage is that we only find patterns that have meaning in the presence of other patterns. This is not necessarily the case in the setting described in the former section since the pattern mining operation itself does not take into account the relationships between patterns, and may produce many patterns which could have been pruned if the pattern search was more aware of the subsequent pattern selection.

A first strategy is to adapt post-processing algorithms. Whereas greedy post-processing algorithms iteratively search for a pattern in a pre-computed set of patterns, this search for patterns can in some cases also be performed by a pattern mining algorithm. The main observation is that given an already selected set of patterns, some scoring functions for measuring the diversity of a new pattern set are boundable, and hence we can use similar strategies to find new patterns as in \pn \ pattern mining \cite{ruckertK07,thoma09}, hence avoiding having to pre-compute a set of patterns.


An alternative approach lies in using a model-dependent iterative strategy (as discussed in Section~\ref{dependent-iterative}); one can ignore the model produced by these strategies afterwards and use the patterns as features in other classification models \cite{cheng08}. The difference between model-dependent and -independent approaches is thus sometimes not as clear-cut as our terminology suggests.

\section{Model-Dependent Pattern Selection}

While all the methods described in the preceding select patterns and sets of 
patterns using scoring functions, these scoring functions are not influenced by 
the choice of model that will be constructed from the patterns. Even though in 
model-independent approaches patterns may be used in SVMs, which show very 
good accuracy and guard against overfitting to a certain degree, the resulting 
models are difficult to interpret. The alternative is to use patterns directly 
to predict class labels, giving users the advantage of being able to examine 
and interpret the model. When doing this, it is often advantageous to adapt the scoring function to the model in which the patterns will be used.
Also here we distinguish the existing methods into post-processing and iterative approaches.


\subsection{\label{dependent-post-processing}Model-dependent post-processing}
\begin{figure}
	\vspace{-5mm}
	\centering
	\includegraphics[width=0.75\textwidth]{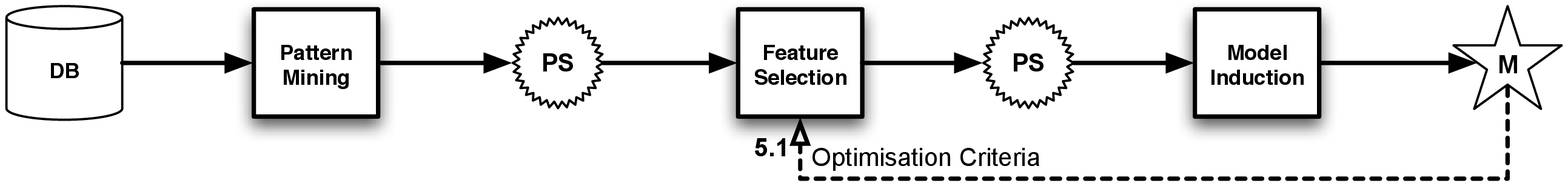}
	\vspace{-5mm}
	\caption{\label{flow_MD-PP}The model-dependent post-processing approach: feedback is given by the model itself to steer feature selection.}
	\vspace{-5mm}
\end{figure}

These approaches are also known as methods for {\em associative classification}. A variety of approaches have been proposed towards building a classifier from rules and performing predictions using selected patterns. 

The simplest such approaches post-process \emph{all} patterns found in a previous phase; they rely on a conflict resolution strategy, similarly to unordered rule lists, which often means that in order to predict which class an example belongs to, a score is computed for each class from the patterns for that class. Many such scoring strategies have been proposed \cite{DBLP:conf/dis/DongZWL98,DBLP:conf/sdm/WangK05,DBLP:journals/kais/LiDR01,xrules,DBLP:conf/www/RamamohanaraoF07,DBLP:conf/kdd/ArunasalamC06,DBLP:conf/pkdd/LeeuwenVS06}. In some cases, such as \cite{DBLP:conf/pkdd/LeeuwenVS06}, another approach on the border between model-dependence and {-~i}ndependence, model-independent pattern selection takes place before patterns are used in such a voting scheme.

The alternative is to perform an ordered heuristic search over a set of patterns, guided by a database coverage constraint. In a sense this is a post-processing version of the sequential covering/weighted covering paradigm known in machine learning. Essentially, these approaches execute these steps:
\begin{enumerate}
 \item they sort the patterns;
 \item they select a pattern according to this sorting order;
 \item they optionally remove some of the remaining unselected patterns;
 \item they optionally resort remaining unselected patterns according to updated scores;
 \item they recursively continue selecting a pattern.
\end{enumerate}
Strategies implementing this idea have been studied in \cite{cba,cmar,ctc}. 

While these approaches construct a model greedily, \cite{nijssen07b} showed that itemsets can be post-processed to construct a decision tree optimally. In this approach, an itemset corresponds to a path from the root to a leaf in a decision tree. Itemsets are selected from a set such that the resulting tree is optimal given user-specified constraints and criteria.

\subsection{\label{dependent-iterative}Model-dependent iterative mining}
\begin{figure}
	\vspace{-5mm}
	\centering
	\includegraphics[width=0.75\textwidth]{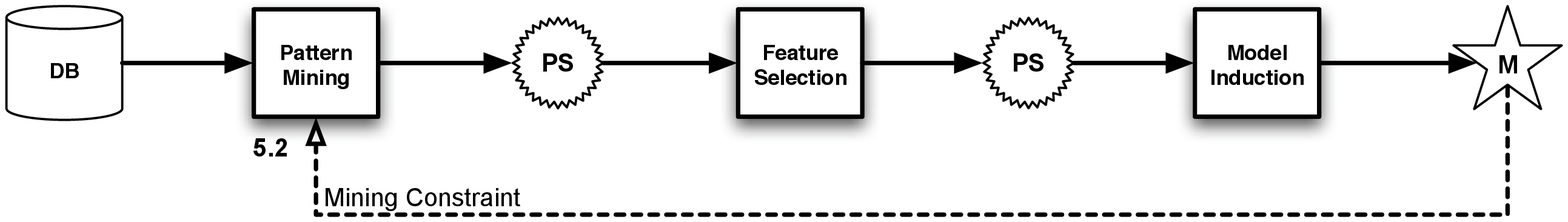}
	\vspace{-5mm}
	\caption{\label{flow_MD-IT}The model-dependent iterative approach: feedback given by the model influences which patterns are mined next.}
	\vspace{-5mm}
\end{figure}
In model-dependent, iterative mining techniques the connection to and inspiration by machine learning becomes most obvious. Hence, these approaches are best understood as adaptations of machine learning techniques. We can distinguish the following classification models.

\begin{description}
\item \emph{FOIL}-like decision list learning strategies: these are techniques that can be understood as adaptations of the FOIL rule learning technique, combined with the weighted covering metaheuristic \cite{DBLP:conf/sdm/YinH03}.

\item \emph{Decision tree} learning strategies: these methods adapt decision tree induction algorithms such as C4.5 \cite{DBLP:conf/pkdd/BringmannZ05,dt-gbi,cheng08icdm}; they iteratively search for \pn\ patterns that split data as well as possible according to criteria such as information gain; the search continues in parallel for the data sets resulting from the split.

\item \emph{Instance-based} learning strategies: these are methods where pattern mining is delayed till a test example is given; \pn\ patterns are searched that are relevant for the test instance \cite{DBLP:conf/icdm/VelosoMZ06,DBLP:conf/pkdd/VelosoMGZ07,DBLP:conf/pkdd/LiDR00}.

\item \emph{Boosting} strategies: these are methods in which classifications of patterns are weighted, and rules are found by iteratively reweighting examples \cite{gboost1,gboost2}.

\item \emph{Regression} strategies: these are methods in which predictions are based on weighted sums of patterns, and weights of patterns are found by linear regression \cite{saigo08}. The boosting and regression methods often include a regularization parameter which needs to be set. In \cite{DBLP:conf/icml/Tsuda07} it was studied how to find the {\em regularization path}, which in this case can be seen as an ordered set of patterns, each prefix of which corresponds to a regression model for one choice of this parameter.
\end{description}

As pointed out, one can also choose to ignore the model constructed by any model-dependent strategy, and use the patterns as features in another type of model. The two categories show therefore different kinds of flexibility: while model-dependent results can be used both directly and as building blocks of another model, they are probably best suited to the model that was used to derive them, differing from the results of model-independent techniques.

\section{Conclusions}
In this paper we presented our perspective on the area of pattern-based classification. Key elements in our perspective are pattern type and data structure independence; instead, we propose to categorize approaches along two dimensions: whether they are model-dependent or model-independent, and whether they are iterative or non-iterative.

For almost any quality measure and mining techniques, both the pattern language and the language in which data are expressed are \emph{not} relevant for the pattern set selection phase as long as there is a well-defined matching operator between the two. Furthermore, almost all techniques for mining \pn{} patterns \emph{themselves} are independent of these aspects as well, with the exception of data structures used. Such data structures, however, typically do not influence the applicability of mining techniques but only their implementation. This means that it is possible to transfer approaches freely between different representations and settings, albeit possibly at a certain cost of efficiency.

Iterative approaches have the advantage of taking the effects of already selected patterns into account by adjusting the scoring function in some way. This allows to focus on interesting areas of the pattern space, pruning subspaces that would have been explored in non-iterative mining, and visiting others that would have been ignored otherwise. The downside to this is that the space of potential solutions is far larger than in the non-iterative case, requiring the adoption of heuristic techniques and less control over, and looser guarantees for the quality of, resulting sets.
Whereas early approaches were often model-dependent post-processing approaches, recent work more focuses on iterative approaches, both model-dependent and -independent. 

The trade-off involved in model-dependence and -independence has been sketched in the preceding section: the agnosticism of model-independence means that resulting sets can be expected to be useful to different kinds of modeling techniques instead of being tailored towards a particular model as in model-dependent solutions. In addition, while predictive models can be used in scoring functions, they do not exhaust the issue and therefore model-independent techniques can use measures that focus on different aspects of pattern relations and may eschew class labels completely. However, results that are produced by such approaches cannot be expected to be useful as direct predictors, making an additional more or less complex modeling step necessary which will probably reduce interpretability. Model-dependent techniques, on the other hand, usually result in models in which the relationship among particular patterns and between patterns and prediction are far more easily accessible. In addition, resulting pattern sets can still be used as input to a different modeling step but might perform worse than pattern sets produced by model-independent approaches.

A major issue in the current state-of-the-art is that so far it is not very clear to what degree the merits and drawbacks that can be derived analytically for different approaches materialize empirically.
In most of the papers that proposed pattern-based classification algorithms, experiments were performed to show the benefits of the approaches. However, these comparisons were (understandably) often limited; they did not exhaustively consider all relevant comparable approaches that derive if one would take
pattern-type independence into account and recognize that graph-based approaches may also be used in simpler pattern domains;
also, the number of data sets in most publications is limited and usually restricted to one data type. A few recent publications have presented more exhaustive experimental comparisons; \cite{DBLP:journals/tkde/DeshpandeKWK05,DBLP:journals/kais/WaleWK08} compared post-processing pattern based classification with kernels and traditional approaches on a large number of molecular data sets, and obtained mixed results. Similarly, \cite{DBLP:conf/sdm/JanssenF09} compared exhaustive rule discovery strategies to greedy ones on UCI data sets. These results are necessary steps into gaining a better insight in the true relative merits of the many pattern-based classification strategies.

\paragraph*{Acknowledgments}
This work was supported by a Postdoc and a project grant from the Research Foundation---Flanders, project ``Principles of Patternset Mining''.

\bibliographystyle{plain}
\bibliography{refs,tutorial}

\end{document}